\documentclass{article}
\usepackage{amsmath,amssymb,graphicx,mlspconf}
\usepackage{pgfplots}
\usepackage{tikz}
\usetikzlibrary{arrows,shapes,positioning,plotmarks}
\usepackage{graphicx}
\usepackage{color}
\usepackage{tikz}
\usepackage{pgf-umlsd}
\usepackage{ifthen}
\usepackage{linegoal}
\usepackage{graphicx}
\usepackage{xcolor}
\usepackage{subcaption}
\usepackage{balance}

\newlength{\figurewidth}
\newlength{\figureheight}

  \newcommand{\eg}{\textit{e.g.}}

  \usepackage{xcolor}

  \graphicspath{{./fig1/}{./img/}}

  \usepackage{xcolor,colortbl}
  
  \definecolor{DarkGray}{gray}{0.85}
  \definecolor{Gray}{gray}{0.90}

  \definecolor{mycolor1}{rgb}{0.00000,0.44700,0.74100}%
  \definecolor{mycolor2}{rgb}{0.85000,0.32500,0.09800}%
  \definecolor{mycolor3}{rgb}{0.92900,0.69400,0.12500}%
  \definecolor{mycolor4}{rgb}{0.49400,0.18400,0.55600}%
  \definecolor{mycolor5}{rgb}{0.46600,0.67400,0.18800}%
  \definecolor{mycolor6}{rgb}{0.30100,0.74500,0.93300}%
  \definecolor{mycolor7}{rgb}{0.3,0.0,0.3}%
  \definecolor{mycolor8}{rgb}{0.0,0.3,0.3}%
  \definecolor{mycolor9}{rgb}{0.3,0.3,0.3}%
  \definecolor{mycolor10}{rgb}{0.9,0.5,0.5}%
  \definecolor{mycolor11}{rgb}{0.5,0.5,0.9}%
  \definecolor{mycolor12}{rgb}{0.5,0.9,0.5}%

  \usepackage{hyperref}
  \hypersetup{
    bookmarks=true,
    pdfstartview={FitH},
    colorlinks=true,
    linkcolor=black,
    citecolor=black,
    filecolor=black,
    urlcolor=black
  }

  \makeatletter
  \g@addto@macro{\UrlBreaks}{\UrlOrds}
  \makeatother

  \usepackage{mathabx}

  \providecommand{\norm}[1]{\|#1\|}

  \usepackage{bm}
  
  \newcommand{\mbf}[1]{\mathbf{#1}}
  \newcommand{\vect}[1]{\mbf{#1}}

\title{Deep Learning Based Speed Estimation for Constraining Strapdown Inertial Navigation on Smartphones}
\name{Santiago Cort\'es \qquad Arno Solin \qquad Juho Kannala}
\address{Aalto University \\ Department of Computer Science \\ Espoo, Finland}
\begin{document}
\maketitle
\begin{abstract}
  Strapdown inertial navigation systems are sensitive to the quality of the data provided by the accelerometer and gyroscope. Low-grade IMUs in handheld smart-devices pose a problem for inertial odometry on these devices. We propose a scheme for constraining the inertial odometry problem by complementing non-linear state estimation by a CNN-based deep-learning model for inferring the momentary speed based on a window of IMU samples. We show the feasibility of the model using a wide range of data from an iPhone, and present proof-of-concept results for how the model can be combined with an inertial navigation system for three-dimensional inertial navigation.
\end{abstract}
\section{Introduction}
\label{sec:intro}

Recently, there has been a growing interest towards implementing inertial navigation systems (INS, \cite{Jekeli:2001,Britting:2010}) on smart mobile devices \cite{Solin+Cortes+Rahtu+Kannala, Yan+Shan+Furukawa:2017, Chen+Lu+Markham+Trigoni:2018}. Indeed, most current smart devices, like phones, tablets and watches are equipped with inertial measurement units (IMU) implemented as microelectromechanical systems (MEMS). These sensors are typically used for orientation estimation and gravitation tracking \cite{Sarkka+Tolvanen+Kannala+Rahtu:2015} or activity classification, but their use for inertial navigation is challenging due to the low quality and high noise levels of the sensors, because errors accumulate rapidly in the double integration process that is required to get position from accelerations. Yet, there would be plenty of applications and use cases, ranging from indoor navigation to augmented reality and robotics, which would benefit from full six degree of freedom inertial navigation using cheap MEMS based sensors.

Due to the above challenges, there are not yet accurate INS based tracking approaches that would be suitable for handheld smart devices in the general case. However, in some constrained use cases, such as the case of foot-mounted IMUs \cite{Foxlin:2005}, inertial navigation may provide good results thanks to the frequent zero-velocity updates that can be detected automatically when the sensor is at rest \cite{Nilsson+Gupta+Handel:2014}. Also, if the IMU can be combined with a video camera, then visual-inertial odometry (VIO) methods can provide good accuracy \cite{Solin+Cortes+Rahtu+Kannala:2018, maplab}. However, in many use cases VIO is not a possible solution since capturing and processing video causes highly increased battery usage. In addition, it requires unobstructed field of view to an environment that has enough distinguishable visual features. Most VIO methods are not robust to full occlusions.

If foot-mounted sensors or cameras are not available, most inertial sensor based approaches for pedestrian dead-reckoning (PDR) rely on step counting \cite{Harle:2013, Xiao+Wen+Markham+Trigoni:2014, Mansur+Makihara+Aqmar+Yagi:2014}. These methods typically assume horizontal 2D motion and are not applicable for wheeled motion or tracking pedestrians in 3D environments or in escalators and elevators. Moreover, if the device is handheld so that the sensor orientation is not fixed with respect to the walking direction, then heading estimation is an additional difficulty besides step detection and even the most robust current systems (\eg\ \cite{Xiao+Wen+Markham+Trigoni:2014}) have limited performance in challenging use cases as shown by \cite{Chen+Lu+Markham+Trigoni:2018}.

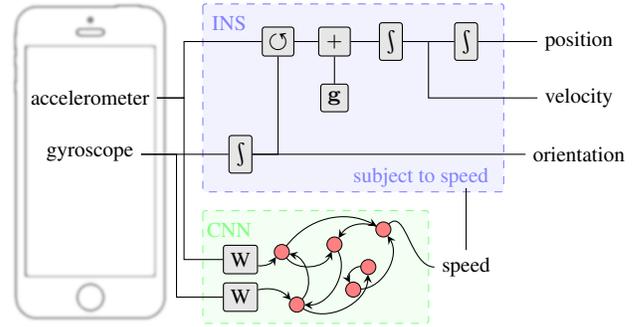
\begin{figure}[t]
  \centering\footnotesize

  \tikzstyle{block} = [draw, node distance=1cm, text badly centered, rounded corners=1pt, fill=black!10]
  \tikzstyle{op} = [draw, node distance=.5cm, text badly centered, rounded corners=1pt, fill=black!10]
  \tikzstyle{branch} = [node distance=.5cm]
  \tikzstyle{node} = [inner sep=0, outer sep=0, minimum size=0]
  \tikzstyle{line} = [draw, postaction={decorate}]
  \tikzstyle{neuron} = [shape=circle,fill=red!50,draw,minimum size=2mm,inner sep=0, outer sep=0,line width=0.01mm]

  \usetikzlibrary{decorations.markings}

  \begin{tikzpicture}[auto]

    \node at (-1.5,-1.6) {\includegraphics[width=2.1cm]{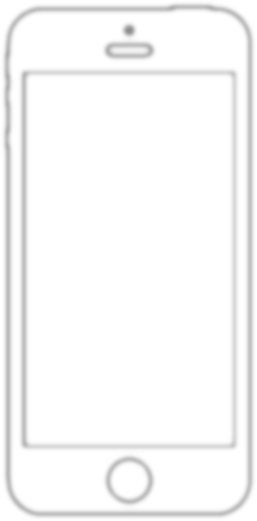}};

    \draw[draw=blue!50,rounded corners=1pt,fill=blue!5,dashed] (0,.5) rectangle ++(4.0,-2.5);
    \draw[draw=green!50,rounded corners=1pt,fill=green!5,dashed] (0,-2.25) rectangle ++(3.0,-1.5);

    \node[branch] at (-1.5,-.75) (acc) {accelerometer};
    \node[block] at (1,0) (acc-rot) {$\circlearrowleft$};
    \node[block] at (1.75,0) (acc-sum) {$+$};
    \node[block] at (1.75,-.75)(grav) {$\vect{g}$};
    \node[block] at (2.5,0) (acc-int) {$\int$};    
    \coordinate (vel) at (3,0);
    \node[op] at (3.5,0) (vel-int) {$\int$};
    \node[branch] at (5,0) (position) {position};
    \node[branch] at (5,-.75)(velocity) {velocity};
    \node[branch] at (5,-1.5) (orientation) {orientation}; 
    \node[branch] at (-1.5,-1.5) (gyro) {gyroscope};
    \node[op] at (.5,-1.5) (gyro-int) {$\int$};

    \node[color=blue!50] at (.35,.25) {INS};
    \node[color=green!50] at (.35,-2.50) {CNN};

    \path [line] (gyro) -- (gyro-int);
    \path [line] (gyro-int) -| (acc-rot); 
    \path [line] (gyro-int) -- (orientation); 
    \path [line] (acc) -- +(1.25,0) |- (acc-rot); 
    \path [line] (acc-rot) -- (acc-sum); 
    \path [line] (grav) -- (acc-sum); 
    \path [line] (acc-sum) -- (acc-int); 
    \path [line] (acc-int) -- (vel); 
    \path [line] (vel) -- (vel-int); 
    \path [line] (vel-int) -- (position); 
    \path [line] (vel) |- (velocity);

    \node[neuron] at (1.25,-3.5) (n1) {};
    \node[neuron] at (1.05,-2.8) (n2) {};
    \node[neuron] at (1.75,-2.7) (n3) {};
    \node[neuron] at (2.20,-3.0) (n4) {};
    \node[neuron] at (2.00,-3.3) (n5) {};
    \node[neuron] at (2.40,-2.5) (n6) {};

    \path [line,->,>=stealth] (n1) to [bend right=60] (n2);
    \path [line,->,>=stealth] (n2) to [bend right=60] (n3);
    \path [line,->,>=stealth] (n3) to [bend left=60] (n1);
    \path [line,->,>=stealth] (n1) to [bend right=60] (n4);
    \path [line,->,>=stealth] (n2) to [bend left=40] (n6);
    \path [line,->,>=stealth] (n6) to [bend right=40] (n3); 
    \path [line,->,>=stealth] (n4) to [bend right=60] (n5);
    \path [line,->,>=stealth] (n5) to [bend right=60] (n6);

    \node[block] at (.5,-2.90) (cnn-input1) {W};
    \node[block] at (.5,-3.40) (cnn-input2) {W};
    \node[branch] at (3.5,-3) (cnn-output) {speed};     
    \path [line] (gyro)-- +(1.15,0) |- (cnn-input2);
    \path [line] (acc)-- +(1.25,0) |- (cnn-input1);
    \path [line,->,>=stealth] (cnn-input1) to [bend right=30] (n2);
    \path [line,->,>=stealth] (cnn-input2) to [bend left=30] (n1);
    \path [line] (n6) to [in=180] (cnn-output);

    \node[color=blue!50] at (2.90,-1.8) {subject to speed};
    \coordinate (st) at (3.5,-1.95);
    \path [line] (cnn-output) -- (st);
    
  \end{tikzpicture}

  \caption{Sketch of our approach. A convolutional network is used to regress speed information which is used to softly constrain an INS system.} 
  \label{fig:sketch}
\end{figure}

In this paper, two different speed concepts are discussed: \emph{Momentary speed} is defined as the average speed over a short window in time. \emph{Instantaneous speed} is defined as the norm of the velocity vector in the state at a given point in time.

Very recently, there have been efforts to utilize machine learning for regressing the momentary horizontal velocity vector from a short time window of gyroscope and accelerometer data \cite{Yan+Shan+Furukawa:2017,Chen+Lu+Markham+Trigoni:2018}. The ground truth motion trajectories for training are obtained with an optical motion capture system (Vicon) \cite{Chen+Lu+Markham+Trigoni:2018} or using visual-inertial odometry (Google Tango) \cite{Yan+Shan+Furukawa:2017}. In \cite{Yan+Shan+Furukawa:2017} velocity is estimated using support vector regression whereas \cite{Chen+Lu+Markham+Trigoni:2018} uses recurrent neural networks (LSTM). Both of the aforementioned approaches train a separate regressor per use case (\eg\ phone in hand, bag, pocket or on trolley) and \cite{Yan+Shan+Furukawa:2017} also trains a support vector classifier which first detects the use case (but is error-prone in practice). Since both approaches only consider horizontal planar motion and performance is not evaluated thoroughly in scenarios where different use cases are mixed (\eg\ phone both in hand and pocket during the same session), it is clear that the existing methods do not yet provide a general and practical six degrees of freedom inertial navigation solution for smart devices.

In this paper we take a different approach and, instead of trying to solve the inertial navigation problem end-to-end using neural networks like \cite{Chen+Lu+Markham+Trigoni:2018}, we aim at combining machine learning based speed estimation with a classical method \cite{Solin+Cortes+Rahtu+Kannala}, which is based on probabilistic sensor fusion using extended Kalman filtering (EKF). That is, we build upon recent work \cite{Solin+Cortes+Rahtu+Kannala}, which has shown that utilizing automatic zero-velocity updates (ZUPTs) and pseudo-measurements for limiting momentary speed can give accurate trajectory estimates in varying use cases. However, often in free handheld movement ZUPTs can not be established frequently enough to constrain motion sufficiently. Hence, in this work we aim at using machine learning to improve the accuracy of the momentary speed estimates which are used as pseudo-measurements in \cite{Solin+Cortes+Rahtu+Kannala}. Thus, instead of assuming a globally constant momentary speed with high uncertainty like in \cite{Solin+Cortes+Rahtu+Kannala}, we predict more accurate data-driven momentary speed estimates that can be used as pseudo-measurements with smaller uncertainty (see Fig.~\ref{fig:sketch}). It should be noted that our approach is not limited to 2D motion like \cite{Yan+Shan+Furukawa:2017,Chen+Lu+Markham+Trigoni:2018}, as we regress the scalar speed and not a 2D velocity vector.

The contributions of this paper are:
\begin{itemize}
  \item A novel neural network model for regressing bounds of speed from short time windows of IMU information, based on recognition of the motion pattern.
  \item Experimentally showing that a single regression model can provide accurate speed estimates in varying use cases involving handheld motion while walking, standing or traveling in elevators and escalators. 
  \item A versatile dataset with visual-inertial odometry based ground truth for training models on inertial only data.
\item Proof-of-concept results showing that the speed estimates provide additional benefits in data-driven inertial navigation for orientation free 3D odometry.
\end{itemize}

This paper is structured as follows. In section~\ref{sec:methods}, we first introduce the problem formulation, then focus on the speed regression subproblem. Then we present the training data collected on a smartphone and go through the pre-processing steps. in section~\ref{sec:experiments}, we present results for the speed regression subproblem (which is the main focus) and then show proof-of-concept results for the constrained INS system in challenging settings. Finally, the model and results are discussed.

\section{Materials and Methods}
\label{sec:methods}
A sketch of our approach is given in Figure~\ref{fig:sketch}. As inputs we use the three-axis gyroscope and accelerometer data from a smartphone. This data is passed to an inertial navigation system (labeled `INS' and adopted following \cite{Solin+Cortes+Rahtu+Kannala}) doing statistical inference on the current 3D position, velocity, and orientation. The blocks `$\int$', `$\circlearrowleft$', and `$\vect{g}$' denote integration, rotation, and gravity, respectively. Our main interest is, however, in the `CNN' block which takes in a window (labeled with the windowing operator `W') of IMU data and tries to infer the momentary scalar speed to be used as a constraint in the INS block.

Given a mode of locomotion, the momentary speed of an agent over a couple of seconds is usually well constrained. In the case of human walking, the momentary speed depends on the person, the terrain and the particular gait. Even in these highly variable cases, there are reasonable bounds for the speed. Those reasonable bounds are introduced into a Kalman filter in the form of a pseudo-velocity update, a weakly informative measurement on the first derivative of the position. 

The raw IMU data contains enough information to calssify the motion into high level gait and transportation classes \cite{Yan+Shan+Furukawa:2017}. Instead of classifying into labeled classes, we directly regress the momentary speed. 

For this we use a small convolutional network that regresses the speed given a two second window of IMU data at 100 Hz. The ground truth velocity is taken as the displacement over the window as reported by ARKit divided by the time difference. This, in theory, is the average of all the instantaneous speeds over the window.

\subsection{Speed regression}

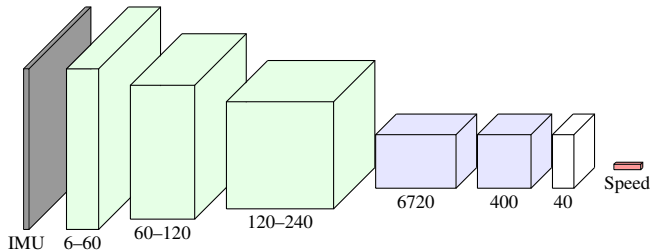
\begin{figure}[t!]
	\centering
	\hspace*{1em}\resizebox{\textwidth}{!}{
	\begin{tikzpicture}
		\draw[use as bounding box, transparent] (-1.8,-1.8) rectangle (23, 3.2);

		\newcommand{\networkLayer}[6]{
			\def\a{#1}
			\def\b{0.02}
			\def\c{#2}
			\def\t{#3}
			\ifthenelse {\equal{#6} {}} {\def\y{0}} {\def\y{#6}}

			\filldraw[#4] (\t+\b,\b,\a) -- (\c+\t-\b,\b,\a) -- (\c+\t-\b,\a-\b,\a) -- (\t+\b,\a-\b,\a) -- (\t+\b,\b,\a);
			\filldraw[#4] (\t+\b,\a,\a-\b) -- (\c+\t-\b,\a,\a-\b) -- (\c+\t-\b,\a,\b) -- (\t+\b,\a,\b);

			\ifthenelse {\equal{#4} {}}
			{}
			{\filldraw[#4] (\c+\t,\b,\a-\b) -- (\c+\t,\b,\b) -- (\c+\t,\a-\b,\b) -- (\c+\t,\a-\b,\a-\b);}

			\draw[line width=0.2mm](\c+\t,0,0) -- (\c+\t,\a,0) -- (\t,\a,0);
			\draw[line width=0.2mm](\t,0,\a) -- (\c+\t,0,\a) node[midway,below] {#5} -- (\c+\t,\a,\a) -- (\t,\a,\a) -- (\t,0,\a);
			\draw[line width=0.2mm](\c+\t,0,0) -- (\c+\t,0,\a);
			\draw[line width=0.2mm](\c+\t,\a,0) -- (\c+\t,\a,\a);
			\draw[line width=0.2mm](\t,\a,0) -- (\t,\a,\a);

		}

		\networkLayer{3.0}{0.1}{-0.8}{color=gray!80}{IMU}

		\networkLayer{3.0}{0.6}{0.0}{color=green!10}{6--60}{}
		\networkLayer{2.5}{1.2}{1.0}{color=green!10}{60--120}{}    
    	\networkLayer{2.0}{2.0}{2.6}{color=green!10}{120--240}{}    

		\networkLayer{1.0}{1.5}{5.0}{color=blue!10}{6720}{}
		\networkLayer{1.0}{1.0}{6.9}{color=blue!10}{400}{}
		\networkLayer{1.0}{0.4}{8.3}{color=white}{40}{}

		\networkLayer{0.1}{0.5}{9.1}{color=red!40}{Speed}{}          

	\end{tikzpicture}
	}
	\caption{Structure of the network used, convolutional layers (green) show the number of input--output channels, while fully connected layers (blue) show the number of units. All layers except the output (white) have rectified linear activation (RELU).}
	\label{fig:cnn}
\end{figure}

The estimation is performed by a convolutional neural network  (see, \eg, \cite{LeCun+Boser:1989}) tasked with regressing the average speed over a finite window of measurements. The hypothesis is that given enough training on a locomotion mode, the momentary speed can be regressed. The input of the network is the six channels of inertial information over a few seconds and the output is the norm of the velocity vector.

The sample windows have a random starting point, which means that the patterns and relations that enable the classification are randomly shifted in the signal. This problem is naturally solved with a convolutional network which is by design invariant to shifts. Once the features are read by the convolutional layers, the fully connected layers perform the regression of the speed.

The structure of the network and the per layer parameters are shown in Figure~\ref{fig:cnn}. All convolutional layers have kernel length 10 and stride of 1. Since the network has a relatively small size, no pooling layers are used. The cost function is the squared error of the momentary speed,
\begin{equation}
  \mathcal{L} = {(S_\mathrm{pred} - \norm{\vect{p}_{T}-\vect{p}_0}/T)}^2,
\end{equation}
where $S_\mathrm{pred}$ is the predicted speed, $\vect{p}_0 \in \mathbb{R}^3$ and $\vect{p}_T \in \mathbb{R}^3$ are respectively the first and last positions of the window and $T$ is the time between the position samples.

The network was trained for 2000 epochs, using mini batches of size 10. Optimization is performed with the Adam algorithm \cite{Kingma+Ba:2014}. It was implemented in pytorch (\url{https://pytorch.org/}) with pandas for data management. Training took roughly seven seconds per epoch in CPU-only training.

\subsection{Data acquisition}
\label{sec:data-acquisition}
For capturing (see details on data acquisition in \cite{Cortes+Solin+Rahtu+Kannala:2018-ADVIO}) authentic sensor data from a smartphone, we implemented (in Swift~4) a data capture application for an Apple iPhone~6s (Apple Inc., \url{https://www.apple.com/iphone-6s}). The application captures raw sensor data from the built-in phone three-axis accelerometer and gyroscope (through the CoreMotion API). Data samples are acquired at 100~Hz and timestamped by the platform. 

For training, we capture phone odometry information from the Apple ARKit (\url{https://developer.apple.com/arkit/}) visual-inertial framework, which runs information fusion of the camera and IMU data for inferring the six degrees-of-freedom relative motion of the device. ARKit can provide a low-drift movement trajectory with associated orientation information. The ARKit output is acquired simultaneously and time-locked to the sensor data output at 60~Hz. Additionally, for reference we also save the video stream of the back-facing iPhone camera at 60~fps.

The training and testing data was captured in various environments by walking around with the mobile phone, primarily indoors on campus and in a shopping mall. The data also features outdoor sequences. The parts of the data featuring movement in stairs and escalators or standing still while holding the device, were manually annotated as a post-processing step (using the reference video stream).

The length of the training data is approximately 100 minutes and it contains some 4.5~km of movement (primarily walking). The example test sequence is a continuous data set of length 391~s / 345~m which includes planar walking, stairs and standstill.

\begin{figure}[!t]
  \raggedright\scriptsize
  \setlength{\figurewidth}{0.47\columnwidth}
  \setlength{\figureheight}{\figurewidth}
  \pgfplotsset{
      compat=newest,    
      hide y axis,
      axis x line*=top,
      tick align=outside,
      axis on top,
      minor tick num=1,
  }
  \hspace*{\fill}  
  \begin{subfigure}[t]{0.3\columnwidth}
    \centering
\begin{tikzpicture}

\begin{axis}[
xmin=-0.1, xmax=2.1,
ymin=0, ymax=909.3,
width=\figurewidth,
height=\figureheight,
tick align=outside,
xtick pos=left,
ymajorticks=false,
x grid style={lightgray!92.026143790849673!black},
y grid style={lightgray!92.026143790849673!black},
legend entries={{Minimum},{Maximum},{Mean}},
legend style={at={(2.0,1.20)},draw=none},
legend cell align={left}
]
\addlegendimage{area legend,fill=violet!60.0!black,draw opacity=0,fill opacity=0.8};
\addlegendimage{area legend,fill=teal!60.0!black,draw opacity=0,fill opacity=0.8};
\addlegendimage{area legend,fill=gray,draw opacity=0,fill opacity=0.8};
\draw[fill=violet!60.0!black,draw opacity=0,fill opacity=0.8] (axis cs:0,0) rectangle (axis cs:0.105263157894737,866);
\draw[fill=violet!60.0!black,draw opacity=0,fill opacity=0.8] (axis cs:0.105263157894737,0) rectangle (axis cs:0.210526315789474,24);
\draw[fill=violet!60.0!black,draw opacity=0,fill opacity=0.8] (axis cs:0.210526315789474,0) rectangle (axis cs:0.315789473684211,19);
\draw[fill=violet!60.0!black,draw opacity=0,fill opacity=0.8] (axis cs:0.315789473684211,0) rectangle (axis cs:0.421052631578947,10);
\draw[fill=violet!60.0!black,draw opacity=0,fill opacity=0.8] (axis cs:0.421052631578947,0) rectangle (axis cs:0.526315789473684,1);
\draw[fill=violet!60.0!black,draw opacity=0,fill opacity=0.8] (axis cs:0.526315789473684,0) rectangle (axis cs:0.631578947368421,3);
\draw[fill=violet!60.0!black,draw opacity=0,fill opacity=0.8] (axis cs:0.631578947368421,0) rectangle (axis cs:0.736842105263158,1);
\draw[fill=violet!60.0!black,draw opacity=0,fill opacity=0.8] (axis cs:0.736842105263158,0) rectangle (axis cs:0.842105263157895,2);
\draw[fill=violet!60.0!black,draw opacity=0,fill opacity=0.8] (axis cs:0.842105263157895,0) rectangle (axis cs:0.947368421052632,1);
\draw[fill=violet!60.0!black,draw opacity=0,fill opacity=0.8] (axis cs:0.947368421052632,0) rectangle (axis cs:1.05263157894737,4);
\draw[fill=violet!60.0!black,draw opacity=0,fill opacity=0.8] (axis cs:1.05263157894737,0) rectangle (axis cs:1.15789473684211,0);
\draw[fill=violet!60.0!black,draw opacity=0,fill opacity=0.8] (axis cs:1.15789473684211,0) rectangle (axis cs:1.26315789473684,2);
\draw[fill=violet!60.0!black,draw opacity=0,fill opacity=0.8] (axis cs:1.26315789473684,0) rectangle (axis cs:1.36842105263158,0);
\draw[fill=violet!60.0!black,draw opacity=0,fill opacity=0.8] (axis cs:1.36842105263158,0) rectangle (axis cs:1.47368421052632,0);
\draw[fill=violet!60.0!black,draw opacity=0,fill opacity=0.8] (axis cs:1.47368421052632,0) rectangle (axis cs:1.57894736842105,0);
\draw[fill=violet!60.0!black,draw opacity=0,fill opacity=0.8] (axis cs:1.57894736842105,0) rectangle (axis cs:1.68421052631579,0);
\draw[fill=violet!60.0!black,draw opacity=0,fill opacity=0.8] (axis cs:1.68421052631579,0) rectangle (axis cs:1.78947368421053,0);
\draw[fill=violet!60.0!black,draw opacity=0,fill opacity=0.8] (axis cs:1.78947368421053,0) rectangle (axis cs:1.89473684210526,0);
\draw[fill=violet!60.0!black,draw opacity=0,fill opacity=0.8] (axis cs:1.89473684210526,0) rectangle (axis cs:2,0);
\draw[fill=teal!60.0!black,draw opacity=0,fill opacity=0.8] (axis cs:0,0) rectangle (axis cs:0.105263157894737,249);
\draw[fill=teal!60.0!black,draw opacity=0,fill opacity=0.8] (axis cs:0.105263157894737,0) rectangle (axis cs:0.210526315789474,270);
\draw[fill=teal!60.0!black,draw opacity=0,fill opacity=0.8] (axis cs:0.210526315789474,0) rectangle (axis cs:0.315789473684211,115);
\draw[fill=teal!60.0!black,draw opacity=0,fill opacity=0.8] (axis cs:0.315789473684211,0) rectangle (axis cs:0.421052631578947,46);
\draw[fill=teal!60.0!black,draw opacity=0,fill opacity=0.8] (axis cs:0.421052631578947,0) rectangle (axis cs:0.526315789473684,62);
\draw[fill=teal!60.0!black,draw opacity=0,fill opacity=0.8] (axis cs:0.526315789473684,0) rectangle (axis cs:0.631578947368421,33);
\draw[fill=teal!60.0!black,draw opacity=0,fill opacity=0.8] (axis cs:0.631578947368421,0) rectangle (axis cs:0.736842105263158,28);
\draw[fill=teal!60.0!black,draw opacity=0,fill opacity=0.8] (axis cs:0.736842105263158,0) rectangle (axis cs:0.842105263157895,18);
\draw[fill=teal!60.0!black,draw opacity=0,fill opacity=0.8] (axis cs:0.842105263157895,0) rectangle (axis cs:0.947368421052632,14);
\draw[fill=teal!60.0!black,draw opacity=0,fill opacity=0.8] (axis cs:0.947368421052632,0) rectangle (axis cs:1.05263157894737,13);
\draw[fill=teal!60.0!black,draw opacity=0,fill opacity=0.8] (axis cs:1.05263157894737,0) rectangle (axis cs:1.15789473684211,3);
\draw[fill=teal!60.0!black,draw opacity=0,fill opacity=0.8] (axis cs:1.15789473684211,0) rectangle (axis cs:1.26315789473684,9);
\draw[fill=teal!60.0!black,draw opacity=0,fill opacity=0.8] (axis cs:1.26315789473684,0) rectangle (axis cs:1.36842105263158,10);
\draw[fill=teal!60.0!black,draw opacity=0,fill opacity=0.8] (axis cs:1.36842105263158,0) rectangle (axis cs:1.47368421052632,7);
\draw[fill=teal!60.0!black,draw opacity=0,fill opacity=0.8] (axis cs:1.47368421052632,0) rectangle (axis cs:1.57894736842105,7);
\draw[fill=teal!60.0!black,draw opacity=0,fill opacity=0.8] (axis cs:1.57894736842105,0) rectangle (axis cs:1.68421052631579,5);
\draw[fill=teal!60.0!black,draw opacity=0,fill opacity=0.8] (axis cs:1.68421052631579,0) rectangle (axis cs:1.78947368421053,5);
\draw[fill=teal!60.0!black,draw opacity=0,fill opacity=0.8] (axis cs:1.78947368421053,0) rectangle (axis cs:1.89473684210526,2);
\draw[fill=teal!60.0!black,draw opacity=0,fill opacity=0.8] (axis cs:1.89473684210526,0) rectangle (axis cs:2,3);
\draw[fill=gray,draw opacity=0,fill opacity=0.8] (axis cs:0,0) rectangle (axis cs:0.105263157894737,762);
\draw[fill=gray,draw opacity=0,fill opacity=0.8] (axis cs:0.105263157894737,0) rectangle (axis cs:0.210526315789474,63);
\draw[fill=gray,draw opacity=0,fill opacity=0.8] (axis cs:0.210526315789474,0) rectangle (axis cs:0.315789473684211,39);
\draw[fill=gray,draw opacity=0,fill opacity=0.8] (axis cs:0.315789473684211,0) rectangle (axis cs:0.421052631578947,16);
\draw[fill=gray,draw opacity=0,fill opacity=0.8] (axis cs:0.421052631578947,0) rectangle (axis cs:0.526315789473684,16);
\draw[fill=gray,draw opacity=0,fill opacity=0.8] (axis cs:0.526315789473684,0) rectangle (axis cs:0.631578947368421,13);
\draw[fill=gray,draw opacity=0,fill opacity=0.8] (axis cs:0.631578947368421,0) rectangle (axis cs:0.736842105263158,1);
\draw[fill=gray,draw opacity=0,fill opacity=0.8] (axis cs:0.736842105263158,0) rectangle (axis cs:0.842105263157895,5);
\draw[fill=gray,draw opacity=0,fill opacity=0.8] (axis cs:0.842105263157895,0) rectangle (axis cs:0.947368421052632,4);
\draw[fill=gray,draw opacity=0,fill opacity=0.8] (axis cs:0.947368421052632,0) rectangle (axis cs:1.05263157894737,2);
\draw[fill=gray,draw opacity=0,fill opacity=0.8] (axis cs:1.05263157894737,0) rectangle (axis cs:1.15789473684211,3);
\draw[fill=gray,draw opacity=0,fill opacity=0.8] (axis cs:1.15789473684211,0) rectangle (axis cs:1.26315789473684,3);
\draw[fill=gray,draw opacity=0,fill opacity=0.8] (axis cs:1.26315789473684,0) rectangle (axis cs:1.36842105263158,6);
\draw[fill=gray,draw opacity=0,fill opacity=0.8] (axis cs:1.36842105263158,0) rectangle (axis cs:1.47368421052632,0);
\draw[fill=gray,draw opacity=0,fill opacity=0.8] (axis cs:1.47368421052632,0) rectangle (axis cs:1.57894736842105,0);
\draw[fill=gray,draw opacity=0,fill opacity=0.8] (axis cs:1.57894736842105,0) rectangle (axis cs:1.68421052631579,0);
\draw[fill=gray,draw opacity=0,fill opacity=0.8] (axis cs:1.68421052631579,0) rectangle (axis cs:1.78947368421053,0);
\draw[fill=gray,draw opacity=0,fill opacity=0.8] (axis cs:1.78947368421053,0) rectangle (axis cs:1.89473684210526,0);
\draw[fill=gray,draw opacity=0,fill opacity=0.8] (axis cs:1.89473684210526,0) rectangle (axis cs:2,0);
\end{axis}

\end{tikzpicture}
    \vspace*{-1em}
    \caption{Static}
    \label{fig:histogram-stat}
  \end{subfigure}%
  \hspace*{\fill}
  \begin{subfigure}[t]{0.3\columnwidth}
    \centering
\begin{tikzpicture}

\begin{axis}[
xmin=-0.1, xmax=2.1,
ymin=0, ymax=771.75,
width=\figurewidth,
height=\figureheight,
tick align=outside,
xtick pos=left,
ymajorticks=false,
x grid style={lightgray!92.026143790849673!black},
y grid style={lightgray!92.026143790849673!black}
]
\addlegendimage{ybar,ybar legend,fill=violet!60.0!black,draw opacity=0,fill opacity=0.8};
\draw[fill=violet!60.0!black,draw opacity=0,fill opacity=0.8] (axis cs:0,0) rectangle (axis cs:0.105263157894737,208);
\draw[fill=violet!60.0!black,draw opacity=0,fill opacity=0.8] (axis cs:0.105263157894737,0) rectangle (axis cs:0.210526315789474,240);
\draw[fill=violet!60.0!black,draw opacity=0,fill opacity=0.8] (axis cs:0.210526315789474,0) rectangle (axis cs:0.315789473684211,301);
\draw[fill=violet!60.0!black,draw opacity=0,fill opacity=0.8] (axis cs:0.315789473684211,0) rectangle (axis cs:0.421052631578947,284);
\draw[fill=violet!60.0!black,draw opacity=0,fill opacity=0.8] (axis cs:0.421052631578947,0) rectangle (axis cs:0.526315789473684,293);
\draw[fill=violet!60.0!black,draw opacity=0,fill opacity=0.8] (axis cs:0.526315789473684,0) rectangle (axis cs:0.631578947368421,337);
\draw[fill=violet!60.0!black,draw opacity=0,fill opacity=0.8] (axis cs:0.631578947368421,0) rectangle (axis cs:0.736842105263158,369);
\draw[fill=violet!60.0!black,draw opacity=0,fill opacity=0.8] (axis cs:0.736842105263158,0) rectangle (axis cs:0.842105263157895,394);
\draw[fill=violet!60.0!black,draw opacity=0,fill opacity=0.8] (axis cs:0.842105263157895,0) rectangle (axis cs:0.947368421052632,301);
\draw[fill=violet!60.0!black,draw opacity=0,fill opacity=0.8] (axis cs:0.947368421052632,0) rectangle (axis cs:1.05263157894737,223);
\draw[fill=violet!60.0!black,draw opacity=0,fill opacity=0.8] (axis cs:1.05263157894737,0) rectangle (axis cs:1.15789473684211,119);
\draw[fill=violet!60.0!black,draw opacity=0,fill opacity=0.8] (axis cs:1.15789473684211,0) rectangle (axis cs:1.26315789473684,55);
\draw[fill=violet!60.0!black,draw opacity=0,fill opacity=0.8] (axis cs:1.26315789473684,0) rectangle (axis cs:1.36842105263158,5);
\draw[fill=violet!60.0!black,draw opacity=0,fill opacity=0.8] (axis cs:1.36842105263158,0) rectangle (axis cs:1.47368421052632,0);
\draw[fill=violet!60.0!black,draw opacity=0,fill opacity=0.8] (axis cs:1.47368421052632,0) rectangle (axis cs:1.57894736842105,0);
\draw[fill=violet!60.0!black,draw opacity=0,fill opacity=0.8] (axis cs:1.57894736842105,0) rectangle (axis cs:1.68421052631579,0);
\draw[fill=violet!60.0!black,draw opacity=0,fill opacity=0.8] (axis cs:1.68421052631579,0) rectangle (axis cs:1.78947368421053,0);
\draw[fill=violet!60.0!black,draw opacity=0,fill opacity=0.8] (axis cs:1.78947368421053,0) rectangle (axis cs:1.89473684210526,0);
\draw[fill=violet!60.0!black,draw opacity=0,fill opacity=0.8] (axis cs:1.89473684210526,0) rectangle (axis cs:2,0);
\draw[fill=teal!60.0!black,draw opacity=0,fill opacity=0.8] (axis cs:0,0) rectangle (axis cs:0.105263157894737,17);
\draw[fill=teal!60.0!black,draw opacity=0,fill opacity=0.8] (axis cs:0.105263157894737,0) rectangle (axis cs:0.210526315789474,6);
\draw[fill=teal!60.0!black,draw opacity=0,fill opacity=0.8] (axis cs:0.210526315789474,0) rectangle (axis cs:0.315789473684211,5);
\draw[fill=teal!60.0!black,draw opacity=0,fill opacity=0.8] (axis cs:0.315789473684211,0) rectangle (axis cs:0.421052631578947,4);
\draw[fill=teal!60.0!black,draw opacity=0,fill opacity=0.8] (axis cs:0.421052631578947,0) rectangle (axis cs:0.526315789473684,9);
\draw[fill=teal!60.0!black,draw opacity=0,fill opacity=0.8] (axis cs:0.526315789473684,0) rectangle (axis cs:0.631578947368421,8);
\draw[fill=teal!60.0!black,draw opacity=0,fill opacity=0.8] (axis cs:0.631578947368421,0) rectangle (axis cs:0.736842105263158,24);
\draw[fill=teal!60.0!black,draw opacity=0,fill opacity=0.8] (axis cs:0.736842105263158,0) rectangle (axis cs:0.842105263157895,29);
\draw[fill=teal!60.0!black,draw opacity=0,fill opacity=0.8] (axis cs:0.842105263157895,0) rectangle (axis cs:0.947368421052632,14);
\draw[fill=teal!60.0!black,draw opacity=0,fill opacity=0.8] (axis cs:0.947368421052632,0) rectangle (axis cs:1.05263157894737,29);
\draw[fill=teal!60.0!black,draw opacity=0,fill opacity=0.8] (axis cs:1.05263157894737,0) rectangle (axis cs:1.15789473684211,54);
\draw[fill=teal!60.0!black,draw opacity=0,fill opacity=0.8] (axis cs:1.15789473684211,0) rectangle (axis cs:1.26315789473684,86);
\draw[fill=teal!60.0!black,draw opacity=0,fill opacity=0.8] (axis cs:1.26315789473684,0) rectangle (axis cs:1.36842105263158,168);
\draw[fill=teal!60.0!black,draw opacity=0,fill opacity=0.8] (axis cs:1.36842105263158,0) rectangle (axis cs:1.47368421052632,287);
\draw[fill=teal!60.0!black,draw opacity=0,fill opacity=0.8] (axis cs:1.47368421052632,0) rectangle (axis cs:1.57894736842105,358);
\draw[fill=teal!60.0!black,draw opacity=0,fill opacity=0.8] (axis cs:1.57894736842105,0) rectangle (axis cs:1.68421052631579,305);
\draw[fill=teal!60.0!black,draw opacity=0,fill opacity=0.8] (axis cs:1.68421052631579,0) rectangle (axis cs:1.78947368421053,335);
\draw[fill=teal!60.0!black,draw opacity=0,fill opacity=0.8] (axis cs:1.78947368421053,0) rectangle (axis cs:1.89473684210526,319);
\draw[fill=teal!60.0!black,draw opacity=0,fill opacity=0.8] (axis cs:1.89473684210526,0) rectangle (axis cs:2,241);
\draw[fill=gray,draw opacity=0,fill opacity=0.8] (axis cs:0,0) rectangle (axis cs:0.105263157894737,55);
\draw[fill=gray,draw opacity=0,fill opacity=0.8] (axis cs:0.105263157894737,0) rectangle (axis cs:0.210526315789474,20);
\draw[fill=gray,draw opacity=0,fill opacity=0.8] (axis cs:0.210526315789474,0) rectangle (axis cs:0.315789473684211,28);
\draw[fill=gray,draw opacity=0,fill opacity=0.8] (axis cs:0.315789473684211,0) rectangle (axis cs:0.421052631578947,32);
\draw[fill=gray,draw opacity=0,fill opacity=0.8] (axis cs:0.421052631578947,0) rectangle (axis cs:0.526315789473684,27);
\draw[fill=gray,draw opacity=0,fill opacity=0.8] (axis cs:0.526315789473684,0) rectangle (axis cs:0.631578947368421,42);
\draw[fill=gray,draw opacity=0,fill opacity=0.8] (axis cs:0.631578947368421,0) rectangle (axis cs:0.736842105263158,49);
\draw[fill=gray,draw opacity=0,fill opacity=0.8] (axis cs:0.736842105263158,0) rectangle (axis cs:0.842105263157895,69);
\draw[fill=gray,draw opacity=0,fill opacity=0.8] (axis cs:0.842105263157895,0) rectangle (axis cs:0.947368421052632,230);
\draw[fill=gray,draw opacity=0,fill opacity=0.8] (axis cs:0.947368421052632,0) rectangle (axis cs:1.05263157894737,450);
\draw[fill=gray,draw opacity=0,fill opacity=0.8] (axis cs:1.05263157894737,0) rectangle (axis cs:1.15789473684211,735);
\draw[fill=gray,draw opacity=0,fill opacity=0.8] (axis cs:1.15789473684211,0) rectangle (axis cs:1.26315789473684,724);
\draw[fill=gray,draw opacity=0,fill opacity=0.8] (axis cs:1.26315789473684,0) rectangle (axis cs:1.36842105263158,453);
\draw[fill=gray,draw opacity=0,fill opacity=0.8] (axis cs:1.36842105263158,0) rectangle (axis cs:1.47368421052632,187);
\draw[fill=gray,draw opacity=0,fill opacity=0.8] (axis cs:1.47368421052632,0) rectangle (axis cs:1.57894736842105,27);
\draw[fill=gray,draw opacity=0,fill opacity=0.8] (axis cs:1.57894736842105,0) rectangle (axis cs:1.68421052631579,1);
\draw[fill=gray,draw opacity=0,fill opacity=0.8] (axis cs:1.68421052631579,0) rectangle (axis cs:1.78947368421053,0);
\draw[fill=gray,draw opacity=0,fill opacity=0.8] (axis cs:1.78947368421053,0) rectangle (axis cs:1.89473684210526,0);
\draw[fill=gray,draw opacity=0,fill opacity=0.8] (axis cs:1.89473684210526,0) rectangle (axis cs:2,0);
\end{axis}

\end{tikzpicture}
    \vspace*{-1em}    
    \caption{Walking}
    \label{fig:histogram-walk}
  \end{subfigure}%
  \hspace*{\fill}
  \\[1em] 
  \begin{subfigure}[t]{0.3\columnwidth}
    \centering
\begin{tikzpicture}

\begin{axis}[
xmin=-0.1, xmax=2.1,
ymin=0, ymax=66.15,
width=\figurewidth,
height=\figureheight,
tick align=outside,
xtick pos=left,
ymajorticks=false,
x grid style={lightgray!92.026143790849673!black},
y grid style={lightgray!92.026143790849673!black}
]
\addlegendimage{ybar,ybar legend,fill=violet!60.0!black,draw opacity=0,fill opacity=0.8};
\draw[fill=violet!60.0!black,draw opacity=0,fill opacity=0.8] (axis cs:0,0) rectangle (axis cs:0.105263157894737,8);
\draw[fill=violet!60.0!black,draw opacity=0,fill opacity=0.8] (axis cs:0.105263157894737,0) rectangle (axis cs:0.210526315789474,27);
\draw[fill=violet!60.0!black,draw opacity=0,fill opacity=0.8] (axis cs:0.210526315789474,0) rectangle (axis cs:0.315789473684211,40);
\draw[fill=violet!60.0!black,draw opacity=0,fill opacity=0.8] (axis cs:0.315789473684211,0) rectangle (axis cs:0.421052631578947,39);
\draw[fill=violet!60.0!black,draw opacity=0,fill opacity=0.8] (axis cs:0.421052631578947,0) rectangle (axis cs:0.526315789473684,35);
\draw[fill=violet!60.0!black,draw opacity=0,fill opacity=0.8] (axis cs:0.526315789473684,0) rectangle (axis cs:0.631578947368421,19);
\draw[fill=violet!60.0!black,draw opacity=0,fill opacity=0.8] (axis cs:0.631578947368421,0) rectangle (axis cs:0.736842105263158,5);
\draw[fill=violet!60.0!black,draw opacity=0,fill opacity=0.8] (axis cs:0.736842105263158,0) rectangle (axis cs:0.842105263157895,3);
\draw[fill=violet!60.0!black,draw opacity=0,fill opacity=0.8] (axis cs:0.842105263157895,0) rectangle (axis cs:0.947368421052632,4);
\draw[fill=violet!60.0!black,draw opacity=0,fill opacity=0.8] (axis cs:0.947368421052632,0) rectangle (axis cs:1.05263157894737,1);
\draw[fill=violet!60.0!black,draw opacity=0,fill opacity=0.8] (axis cs:1.05263157894737,0) rectangle (axis cs:1.15789473684211,0);
\draw[fill=violet!60.0!black,draw opacity=0,fill opacity=0.8] (axis cs:1.15789473684211,0) rectangle (axis cs:1.26315789473684,0);
\draw[fill=violet!60.0!black,draw opacity=0,fill opacity=0.8] (axis cs:1.26315789473684,0) rectangle (axis cs:1.36842105263158,0);
\draw[fill=violet!60.0!black,draw opacity=0,fill opacity=0.8] (axis cs:1.36842105263158,0) rectangle (axis cs:1.47368421052632,0);
\draw[fill=violet!60.0!black,draw opacity=0,fill opacity=0.8] (axis cs:1.47368421052632,0) rectangle (axis cs:1.57894736842105,0);
\draw[fill=violet!60.0!black,draw opacity=0,fill opacity=0.8] (axis cs:1.57894736842105,0) rectangle (axis cs:1.68421052631579,0);
\draw[fill=violet!60.0!black,draw opacity=0,fill opacity=0.8] (axis cs:1.68421052631579,0) rectangle (axis cs:1.78947368421053,0);
\draw[fill=violet!60.0!black,draw opacity=0,fill opacity=0.8] (axis cs:1.78947368421053,0) rectangle (axis cs:1.89473684210526,0);
\draw[fill=violet!60.0!black,draw opacity=0,fill opacity=0.8] (axis cs:1.89473684210526,0) rectangle (axis cs:2,0);
\draw[fill=teal!60.0!black,draw opacity=0,fill opacity=0.8] (axis cs:0,0) rectangle (axis cs:0.105263157894737,0);
\draw[fill=teal!60.0!black,draw opacity=0,fill opacity=0.8] (axis cs:0.105263157894737,0) rectangle (axis cs:0.210526315789474,0);
\draw[fill=teal!60.0!black,draw opacity=0,fill opacity=0.8] (axis cs:0.210526315789474,0) rectangle (axis cs:0.315789473684211,0);
\draw[fill=teal!60.0!black,draw opacity=0,fill opacity=0.8] (axis cs:0.315789473684211,0) rectangle (axis cs:0.421052631578947,0);
\draw[fill=teal!60.0!black,draw opacity=0,fill opacity=0.8] (axis cs:0.421052631578947,0) rectangle (axis cs:0.526315789473684,0);
\draw[fill=teal!60.0!black,draw opacity=0,fill opacity=0.8] (axis cs:0.526315789473684,0) rectangle (axis cs:0.631578947368421,1);
\draw[fill=teal!60.0!black,draw opacity=0,fill opacity=0.8] (axis cs:0.631578947368421,0) rectangle (axis cs:0.736842105263158,2);
\draw[fill=teal!60.0!black,draw opacity=0,fill opacity=0.8] (axis cs:0.736842105263158,0) rectangle (axis cs:0.842105263157895,10);
\draw[fill=teal!60.0!black,draw opacity=0,fill opacity=0.8] (axis cs:0.842105263157895,0) rectangle (axis cs:0.947368421052632,23);
\draw[fill=teal!60.0!black,draw opacity=0,fill opacity=0.8] (axis cs:0.947368421052632,0) rectangle (axis cs:1.05263157894737,13);
\draw[fill=teal!60.0!black,draw opacity=0,fill opacity=0.8] (axis cs:1.05263157894737,0) rectangle (axis cs:1.15789473684211,17);
\draw[fill=teal!60.0!black,draw opacity=0,fill opacity=0.8] (axis cs:1.15789473684211,0) rectangle (axis cs:1.26315789473684,24);
\draw[fill=teal!60.0!black,draw opacity=0,fill opacity=0.8] (axis cs:1.26315789473684,0) rectangle (axis cs:1.36842105263158,11);
\draw[fill=teal!60.0!black,draw opacity=0,fill opacity=0.8] (axis cs:1.36842105263158,0) rectangle (axis cs:1.47368421052632,23);
\draw[fill=teal!60.0!black,draw opacity=0,fill opacity=0.8] (axis cs:1.47368421052632,0) rectangle (axis cs:1.57894736842105,13);
\draw[fill=teal!60.0!black,draw opacity=0,fill opacity=0.8] (axis cs:1.57894736842105,0) rectangle (axis cs:1.68421052631579,15);
\draw[fill=teal!60.0!black,draw opacity=0,fill opacity=0.8] (axis cs:1.68421052631579,0) rectangle (axis cs:1.78947368421053,7);
\draw[fill=teal!60.0!black,draw opacity=0,fill opacity=0.8] (axis cs:1.78947368421053,0) rectangle (axis cs:1.89473684210526,5);
\draw[fill=teal!60.0!black,draw opacity=0,fill opacity=0.8] (axis cs:1.89473684210526,0) rectangle (axis cs:2,6);
\draw[fill=gray,draw opacity=0,fill opacity=0.8] (axis cs:0,0) rectangle (axis cs:0.105263157894737,3);
\draw[fill=gray,draw opacity=0,fill opacity=0.8] (axis cs:0.105263157894737,0) rectangle (axis cs:0.210526315789474,0);
\draw[fill=gray,draw opacity=0,fill opacity=0.8] (axis cs:0.210526315789474,0) rectangle (axis cs:0.315789473684211,1);
\draw[fill=gray,draw opacity=0,fill opacity=0.8] (axis cs:0.315789473684211,0) rectangle (axis cs:0.421052631578947,3);
\draw[fill=gray,draw opacity=0,fill opacity=0.8] (axis cs:0.421052631578947,0) rectangle (axis cs:0.526315789473684,7);
\draw[fill=gray,draw opacity=0,fill opacity=0.8] (axis cs:0.526315789473684,0) rectangle (axis cs:0.631578947368421,50);
\draw[fill=gray,draw opacity=0,fill opacity=0.8] (axis cs:0.631578947368421,0) rectangle (axis cs:0.736842105263158,63);
\draw[fill=gray,draw opacity=0,fill opacity=0.8] (axis cs:0.736842105263158,0) rectangle (axis cs:0.842105263157895,27);
\draw[fill=gray,draw opacity=0,fill opacity=0.8] (axis cs:0.842105263157895,0) rectangle (axis cs:0.947368421052632,4);
\draw[fill=gray,draw opacity=0,fill opacity=0.8] (axis cs:0.947368421052632,0) rectangle (axis cs:1.05263157894737,8);
\draw[fill=gray,draw opacity=0,fill opacity=0.8] (axis cs:1.05263157894737,0) rectangle (axis cs:1.15789473684211,12);
\draw[fill=gray,draw opacity=0,fill opacity=0.8] (axis cs:1.15789473684211,0) rectangle (axis cs:1.26315789473684,2);
\draw[fill=gray,draw opacity=0,fill opacity=0.8] (axis cs:1.26315789473684,0) rectangle (axis cs:1.36842105263158,1);
\draw[fill=gray,draw opacity=0,fill opacity=0.8] (axis cs:1.36842105263158,0) rectangle (axis cs:1.47368421052632,0);
\draw[fill=gray,draw opacity=0,fill opacity=0.8] (axis cs:1.47368421052632,0) rectangle (axis cs:1.57894736842105,0);
\draw[fill=gray,draw opacity=0,fill opacity=0.8] (axis cs:1.57894736842105,0) rectangle (axis cs:1.68421052631579,0);
\draw[fill=gray,draw opacity=0,fill opacity=0.8] (axis cs:1.68421052631579,0) rectangle (axis cs:1.78947368421053,0);
\draw[fill=gray,draw opacity=0,fill opacity=0.8] (axis cs:1.78947368421053,0) rectangle (axis cs:1.89473684210526,0);
\draw[fill=gray,draw opacity=0,fill opacity=0.8] (axis cs:1.89473684210526,0) rectangle (axis cs:2,0);
\end{axis}

\end{tikzpicture}
    \vspace*{-1em}    
    \caption{Stairs}
    \label{fig:histogram-stairs}
  \end{subfigure}
  \hspace*{\fill}
  \begin{subfigure}[t]{0.3\columnwidth}
    \centering
\begin{tikzpicture}
  
\begin{axis}[
xmin=-0.1, xmax=2.1,
ymin=0, ymax=16.8,
width=\figurewidth,
height=\figureheight,
tick align=outside,
xtick pos=left,
ymajorticks=false,
x grid style={lightgray!92.026143790849673!black},
y grid style={lightgray!92.026143790849673!black}
]
\addlegendimage{ybar,ybar legend,fill=violet!60.0!black,draw opacity=0,fill opacity=0.8};
\draw[fill=violet!60.0!black,draw opacity=0,fill opacity=0.8] (axis cs:0,0) rectangle (axis cs:0.105263157894737,16);
\draw[fill=violet!60.0!black,draw opacity=0,fill opacity=0.8] (axis cs:0.105263157894737,0) rectangle (axis cs:0.210526315789474,0);
\draw[fill=violet!60.0!black,draw opacity=0,fill opacity=0.8] (axis cs:0.210526315789474,0) rectangle (axis cs:0.315789473684211,3);
\draw[fill=violet!60.0!black,draw opacity=0,fill opacity=0.8] (axis cs:0.315789473684211,0) rectangle (axis cs:0.421052631578947,0);
\draw[fill=violet!60.0!black,draw opacity=0,fill opacity=0.8] (axis cs:0.421052631578947,0) rectangle (axis cs:0.526315789473684,0);
\draw[fill=violet!60.0!black,draw opacity=0,fill opacity=0.8] (axis cs:0.526315789473684,0) rectangle (axis cs:0.631578947368421,0);
\draw[fill=violet!60.0!black,draw opacity=0,fill opacity=0.8] (axis cs:0.631578947368421,0) rectangle (axis cs:0.736842105263158,0);
\draw[fill=violet!60.0!black,draw opacity=0,fill opacity=0.8] (axis cs:0.736842105263158,0) rectangle (axis cs:0.842105263157895,0);
\draw[fill=violet!60.0!black,draw opacity=0,fill opacity=0.8] (axis cs:0.842105263157895,0) rectangle (axis cs:0.947368421052632,1);
\draw[fill=violet!60.0!black,draw opacity=0,fill opacity=0.8] (axis cs:0.947368421052632,0) rectangle (axis cs:1.05263157894737,0);
\draw[fill=violet!60.0!black,draw opacity=0,fill opacity=0.8] (axis cs:1.05263157894737,0) rectangle (axis cs:1.15789473684211,0);
\draw[fill=violet!60.0!black,draw opacity=0,fill opacity=0.8] (axis cs:1.15789473684211,0) rectangle (axis cs:1.26315789473684,0);
\draw[fill=violet!60.0!black,draw opacity=0,fill opacity=0.8] (axis cs:1.26315789473684,0) rectangle (axis cs:1.36842105263158,0);
\draw[fill=violet!60.0!black,draw opacity=0,fill opacity=0.8] (axis cs:1.36842105263158,0) rectangle (axis cs:1.47368421052632,0);
\draw[fill=violet!60.0!black,draw opacity=0,fill opacity=0.8] (axis cs:1.47368421052632,0) rectangle (axis cs:1.57894736842105,0);
\draw[fill=violet!60.0!black,draw opacity=0,fill opacity=0.8] (axis cs:1.57894736842105,0) rectangle (axis cs:1.68421052631579,0);
\draw[fill=violet!60.0!black,draw opacity=0,fill opacity=0.8] (axis cs:1.68421052631579,0) rectangle (axis cs:1.78947368421053,0);
\draw[fill=violet!60.0!black,draw opacity=0,fill opacity=0.8] (axis cs:1.78947368421053,0) rectangle (axis cs:1.89473684210526,0);
\draw[fill=violet!60.0!black,draw opacity=0,fill opacity=0.8] (axis cs:1.89473684210526,0) rectangle (axis cs:2,0);
\draw[fill=teal!60.0!black,draw opacity=0,fill opacity=0.8] (axis cs:0,0) rectangle (axis cs:0.105263157894737,1);
\draw[fill=teal!60.0!black,draw opacity=0,fill opacity=0.8] (axis cs:0.105263157894737,0) rectangle (axis cs:0.210526315789474,3);
\draw[fill=teal!60.0!black,draw opacity=0,fill opacity=0.8] (axis cs:0.210526315789474,0) rectangle (axis cs:0.315789473684211,3);
\draw[fill=teal!60.0!black,draw opacity=0,fill opacity=0.8] (axis cs:0.315789473684211,0) rectangle (axis cs:0.421052631578947,0);
\draw[fill=teal!60.0!black,draw opacity=0,fill opacity=0.8] (axis cs:0.421052631578947,0) rectangle (axis cs:0.526315789473684,5);
\draw[fill=teal!60.0!black,draw opacity=0,fill opacity=0.8] (axis cs:0.526315789473684,0) rectangle (axis cs:0.631578947368421,0);
\draw[fill=teal!60.0!black,draw opacity=0,fill opacity=0.8] (axis cs:0.631578947368421,0) rectangle (axis cs:0.736842105263158,1);
\draw[fill=teal!60.0!black,draw opacity=0,fill opacity=0.8] (axis cs:0.736842105263158,0) rectangle (axis cs:0.842105263157895,0);
\draw[fill=teal!60.0!black,draw opacity=0,fill opacity=0.8] (axis cs:0.842105263157895,0) rectangle (axis cs:0.947368421052632,3);
\draw[fill=teal!60.0!black,draw opacity=0,fill opacity=0.8] (axis cs:0.947368421052632,0) rectangle (axis cs:1.05263157894737,2);
\draw[fill=teal!60.0!black,draw opacity=0,fill opacity=0.8] (axis cs:1.05263157894737,0) rectangle (axis cs:1.15789473684211,2);
\draw[fill=teal!60.0!black,draw opacity=0,fill opacity=0.8] (axis cs:1.15789473684211,0) rectangle (axis cs:1.26315789473684,0);
\draw[fill=teal!60.0!black,draw opacity=0,fill opacity=0.8] (axis cs:1.26315789473684,0) rectangle (axis cs:1.36842105263158,0);
\draw[fill=teal!60.0!black,draw opacity=0,fill opacity=0.8] (axis cs:1.36842105263158,0) rectangle (axis cs:1.47368421052632,0);
\draw[fill=teal!60.0!black,draw opacity=0,fill opacity=0.8] (axis cs:1.47368421052632,0) rectangle (axis cs:1.57894736842105,0);
\draw[fill=teal!60.0!black,draw opacity=0,fill opacity=0.8] (axis cs:1.57894736842105,0) rectangle (axis cs:1.68421052631579,0);
\draw[fill=teal!60.0!black,draw opacity=0,fill opacity=0.8] (axis cs:1.68421052631579,0) rectangle (axis cs:1.78947368421053,0);
\draw[fill=teal!60.0!black,draw opacity=0,fill opacity=0.8] (axis cs:1.78947368421053,0) rectangle (axis cs:1.89473684210526,0);
\draw[fill=teal!60.0!black,draw opacity=0,fill opacity=0.8] (axis cs:1.89473684210526,0) rectangle (axis cs:2,0);
\draw[fill=gray,draw opacity=0,fill opacity=0.8] (axis cs:0,0) rectangle (axis cs:0.105263157894737,10);
\draw[fill=gray,draw opacity=0,fill opacity=0.8] (axis cs:0.105263157894737,0) rectangle (axis cs:0.210526315789474,3);
\draw[fill=gray,draw opacity=0,fill opacity=0.8] (axis cs:0.210526315789474,0) rectangle (axis cs:0.315789473684211,0);
\draw[fill=gray,draw opacity=0,fill opacity=0.8] (axis cs:0.315789473684211,0) rectangle (axis cs:0.421052631578947,2);
\draw[fill=gray,draw opacity=0,fill opacity=0.8] (axis cs:0.421052631578947,0) rectangle (axis cs:0.526315789473684,0);
\draw[fill=gray,draw opacity=0,fill opacity=0.8] (axis cs:0.526315789473684,0) rectangle (axis cs:0.631578947368421,1);
\draw[fill=gray,draw opacity=0,fill opacity=0.8] (axis cs:0.631578947368421,0) rectangle (axis cs:0.736842105263158,2);
\draw[fill=gray,draw opacity=0,fill opacity=0.8] (axis cs:0.736842105263158,0) rectangle (axis cs:0.842105263157895,1);
\draw[fill=gray,draw opacity=0,fill opacity=0.8] (axis cs:0.842105263157895,0) rectangle (axis cs:0.947368421052632,0);
\draw[fill=gray,draw opacity=0,fill opacity=0.8] (axis cs:0.947368421052632,0) rectangle (axis cs:1.05263157894737,1);
\draw[fill=gray,draw opacity=0,fill opacity=0.8] (axis cs:1.05263157894737,0) rectangle (axis cs:1.15789473684211,0);
\draw[fill=gray,draw opacity=0,fill opacity=0.8] (axis cs:1.15789473684211,0) rectangle (axis cs:1.26315789473684,0);
\draw[fill=gray,draw opacity=0,fill opacity=0.8] (axis cs:1.26315789473684,0) rectangle (axis cs:1.36842105263158,0);
\draw[fill=gray,draw opacity=0,fill opacity=0.8] (axis cs:1.36842105263158,0) rectangle (axis cs:1.47368421052632,0);
\draw[fill=gray,draw opacity=0,fill opacity=0.8] (axis cs:1.47368421052632,0) rectangle (axis cs:1.57894736842105,0);
\draw[fill=gray,draw opacity=0,fill opacity=0.8] (axis cs:1.57894736842105,0) rectangle (axis cs:1.68421052631579,0);
\draw[fill=gray,draw opacity=0,fill opacity=0.8] (axis cs:1.68421052631579,0) rectangle (axis cs:1.78947368421053,0);
\draw[fill=gray,draw opacity=0,fill opacity=0.8] (axis cs:1.78947368421053,0) rectangle (axis cs:1.89473684210526,0);
\draw[fill=gray,draw opacity=0,fill opacity=0.8] (axis cs:1.89473684210526,0) rectangle (axis cs:2,0);
\end{axis}

\end{tikzpicture}
    \vspace*{-1em}    
    \caption{Elevator}
    \label{fig:histogram-ele}
  \end{subfigure}%
  \hspace*{\fill}
  \begin{subfigure}[t]{0.3\columnwidth}
    \centering
\begin{tikzpicture}

\begin{axis}[
xmin=-0.1, xmax=2.1,
ymin=0, ymax=46.2,
width=\figurewidth,
height=\figureheight,
tick align=outside,
xtick pos=left,
ymajorticks=false,
x grid style={lightgray!92.026143790849673!black},
y grid style={lightgray!92.026143790849673!black}
]
\addlegendimage{ybar,ybar legend,fill=violet!60.0!black,draw opacity=0,fill opacity=0.8};
\draw[fill=violet!60.0!black,draw opacity=0,fill opacity=0.8] (axis cs:0,0) rectangle (axis cs:0.105263157894737,44);
\draw[fill=violet!60.0!black,draw opacity=0,fill opacity=0.8] (axis cs:0.105263157894737,0) rectangle (axis cs:0.210526315789474,33);
\draw[fill=violet!60.0!black,draw opacity=0,fill opacity=0.8] (axis cs:0.210526315789474,0) rectangle (axis cs:0.315789473684211,27);
\draw[fill=violet!60.0!black,draw opacity=0,fill opacity=0.8] (axis cs:0.315789473684211,0) rectangle (axis cs:0.421052631578947,29);
\draw[fill=violet!60.0!black,draw opacity=0,fill opacity=0.8] (axis cs:0.421052631578947,0) rectangle (axis cs:0.526315789473684,5);
\draw[fill=violet!60.0!black,draw opacity=0,fill opacity=0.8] (axis cs:0.526315789473684,0) rectangle (axis cs:0.631578947368421,4);
\draw[fill=violet!60.0!black,draw opacity=0,fill opacity=0.8] (axis cs:0.631578947368421,0) rectangle (axis cs:0.736842105263158,4);
\draw[fill=violet!60.0!black,draw opacity=0,fill opacity=0.8] (axis cs:0.736842105263158,0) rectangle (axis cs:0.842105263157895,10);
\draw[fill=violet!60.0!black,draw opacity=0,fill opacity=0.8] (axis cs:0.842105263157895,0) rectangle (axis cs:0.947368421052632,1);
\draw[fill=violet!60.0!black,draw opacity=0,fill opacity=0.8] (axis cs:0.947368421052632,0) rectangle (axis cs:1.05263157894737,1);
\draw[fill=violet!60.0!black,draw opacity=0,fill opacity=0.8] (axis cs:1.05263157894737,0) rectangle (axis cs:1.15789473684211,0);
\draw[fill=violet!60.0!black,draw opacity=0,fill opacity=0.8] (axis cs:1.15789473684211,0) rectangle (axis cs:1.26315789473684,0);
\draw[fill=violet!60.0!black,draw opacity=0,fill opacity=0.8] (axis cs:1.26315789473684,0) rectangle (axis cs:1.36842105263158,0);
\draw[fill=violet!60.0!black,draw opacity=0,fill opacity=0.8] (axis cs:1.36842105263158,0) rectangle (axis cs:1.47368421052632,0);
\draw[fill=violet!60.0!black,draw opacity=0,fill opacity=0.8] (axis cs:1.47368421052632,0) rectangle (axis cs:1.57894736842105,0);
\draw[fill=violet!60.0!black,draw opacity=0,fill opacity=0.8] (axis cs:1.57894736842105,0) rectangle (axis cs:1.68421052631579,0);
\draw[fill=violet!60.0!black,draw opacity=0,fill opacity=0.8] (axis cs:1.68421052631579,0) rectangle (axis cs:1.78947368421053,0);
\draw[fill=violet!60.0!black,draw opacity=0,fill opacity=0.8] (axis cs:1.78947368421053,0) rectangle (axis cs:1.89473684210526,0);
\draw[fill=violet!60.0!black,draw opacity=0,fill opacity=0.8] (axis cs:1.89473684210526,0) rectangle (axis cs:2,0);
\draw[fill=teal!60.0!black,draw opacity=0,fill opacity=0.8] (axis cs:0,0) rectangle (axis cs:0.105263157894737,0);
\draw[fill=teal!60.0!black,draw opacity=0,fill opacity=0.8] (axis cs:0.105263157894737,0) rectangle (axis cs:0.210526315789474,0);
\draw[fill=teal!60.0!black,draw opacity=0,fill opacity=0.8] (axis cs:0.210526315789474,0) rectangle (axis cs:0.315789473684211,1);
\draw[fill=teal!60.0!black,draw opacity=0,fill opacity=0.8] (axis cs:0.315789473684211,0) rectangle (axis cs:0.421052631578947,5);
\draw[fill=teal!60.0!black,draw opacity=0,fill opacity=0.8] (axis cs:0.421052631578947,0) rectangle (axis cs:0.526315789473684,6);
\draw[fill=teal!60.0!black,draw opacity=0,fill opacity=0.8] (axis cs:0.526315789473684,0) rectangle (axis cs:0.631578947368421,10);
\draw[fill=teal!60.0!black,draw opacity=0,fill opacity=0.8] (axis cs:0.631578947368421,0) rectangle (axis cs:0.736842105263158,17);
\draw[fill=teal!60.0!black,draw opacity=0,fill opacity=0.8] (axis cs:0.736842105263158,0) rectangle (axis cs:0.842105263157895,14);
\draw[fill=teal!60.0!black,draw opacity=0,fill opacity=0.8] (axis cs:0.842105263157895,0) rectangle (axis cs:0.947368421052632,7);
\draw[fill=teal!60.0!black,draw opacity=0,fill opacity=0.8] (axis cs:0.947368421052632,0) rectangle (axis cs:1.05263157894737,9);
\draw[fill=teal!60.0!black,draw opacity=0,fill opacity=0.8] (axis cs:1.05263157894737,0) rectangle (axis cs:1.15789473684211,9);
\draw[fill=teal!60.0!black,draw opacity=0,fill opacity=0.8] (axis cs:1.15789473684211,0) rectangle (axis cs:1.26315789473684,13);
\draw[fill=teal!60.0!black,draw opacity=0,fill opacity=0.8] (axis cs:1.26315789473684,0) rectangle (axis cs:1.36842105263158,9);
\draw[fill=teal!60.0!black,draw opacity=0,fill opacity=0.8] (axis cs:1.36842105263158,0) rectangle (axis cs:1.47368421052632,12);
\draw[fill=teal!60.0!black,draw opacity=0,fill opacity=0.8] (axis cs:1.47368421052632,0) rectangle (axis cs:1.57894736842105,9);
\draw[fill=teal!60.0!black,draw opacity=0,fill opacity=0.8] (axis cs:1.57894736842105,0) rectangle (axis cs:1.68421052631579,7);
\draw[fill=teal!60.0!black,draw opacity=0,fill opacity=0.8] (axis cs:1.68421052631579,0) rectangle (axis cs:1.78947368421053,4);
\draw[fill=teal!60.0!black,draw opacity=0,fill opacity=0.8] (axis cs:1.78947368421053,0) rectangle (axis cs:1.89473684210526,4);
\draw[fill=teal!60.0!black,draw opacity=0,fill opacity=0.8] (axis cs:1.89473684210526,0) rectangle (axis cs:2,1);
\draw[fill=gray,draw opacity=0,fill opacity=0.8] (axis cs:0,0) rectangle (axis cs:0.105263157894737,12);
\draw[fill=gray,draw opacity=0,fill opacity=0.8] (axis cs:0.105263157894737,0) rectangle (axis cs:0.210526315789474,14);
\draw[fill=gray,draw opacity=0,fill opacity=0.8] (axis cs:0.210526315789474,0) rectangle (axis cs:0.315789473684211,12);
\draw[fill=gray,draw opacity=0,fill opacity=0.8] (axis cs:0.315789473684211,0) rectangle (axis cs:0.421052631578947,31);
\draw[fill=gray,draw opacity=0,fill opacity=0.8] (axis cs:0.421052631578947,0) rectangle (axis cs:0.526315789473684,39);
\draw[fill=gray,draw opacity=0,fill opacity=0.8] (axis cs:0.526315789473684,0) rectangle (axis cs:0.631578947368421,14);
\draw[fill=gray,draw opacity=0,fill opacity=0.8] (axis cs:0.631578947368421,0) rectangle (axis cs:0.736842105263158,5);
\draw[fill=gray,draw opacity=0,fill opacity=0.8] (axis cs:0.736842105263158,0) rectangle (axis cs:0.842105263157895,9);
\draw[fill=gray,draw opacity=0,fill opacity=0.8] (axis cs:0.842105263157895,0) rectangle (axis cs:0.947368421052632,11);
\draw[fill=gray,draw opacity=0,fill opacity=0.8] (axis cs:0.947368421052632,0) rectangle (axis cs:1.05263157894737,7);
\draw[fill=gray,draw opacity=0,fill opacity=0.8] (axis cs:1.05263157894737,0) rectangle (axis cs:1.15789473684211,4);
\draw[fill=gray,draw opacity=0,fill opacity=0.8] (axis cs:1.15789473684211,0) rectangle (axis cs:1.26315789473684,0);
\draw[fill=gray,draw opacity=0,fill opacity=0.8] (axis cs:1.26315789473684,0) rectangle (axis cs:1.36842105263158,0);
\draw[fill=gray,draw opacity=0,fill opacity=0.8] (axis cs:1.36842105263158,0) rectangle (axis cs:1.47368421052632,0);
\draw[fill=gray,draw opacity=0,fill opacity=0.8] (axis cs:1.47368421052632,0) rectangle (axis cs:1.57894736842105,0);
\draw[fill=gray,draw opacity=0,fill opacity=0.8] (axis cs:1.57894736842105,0) rectangle (axis cs:1.68421052631579,0);
\draw[fill=gray,draw opacity=0,fill opacity=0.8] (axis cs:1.68421052631579,0) rectangle (axis cs:1.78947368421053,0);
\draw[fill=gray,draw opacity=0,fill opacity=0.8] (axis cs:1.78947368421053,0) rectangle (axis cs:1.89473684210526,0);
\draw[fill=gray,draw opacity=0,fill opacity=0.8] (axis cs:1.89473684210526,0) rectangle (axis cs:2,0);
\end{axis}

\end{tikzpicture}
    \vspace*{-1em}    
    \caption{Escalator}
    \label{fig:histogram-esc}
  \end{subfigure}
  \caption{Histograms of \textcolor{mycolor7}{\textbf{minimum}}, \textcolor{mycolor8}{\textbf{maximum}}, and \textcolor{mycolor9}{\textbf{mean}} instantaneous speed (m/s) for each window in the labeled modes of locomotion.}
  \label{fig:histograms}
\end{figure}
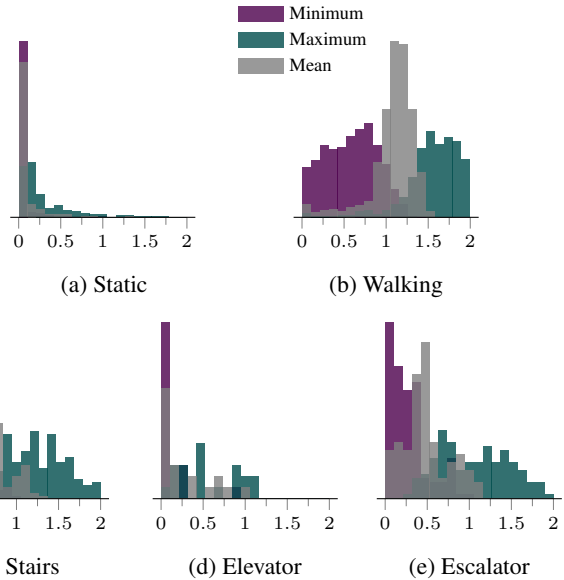

\subsection{Pre-processing}

The IMU measurements are in the phone coordinate frame, the ARKit position is in a global coordinate frame. For the scalar speed, this discrepancy is not a problem, but for more sophisticated measurements, the discrepancy should be resolved. The orientation of the ARKit coordinate frame can be estimated from the gravity vector.

For training, we perform 10-fold cross-validation, where the data is randomly split into 90\% training and 10\% validation sets. A separate test sequence is held out and not used in training. 

The sequences contain normal walking, standing still, and movement in staircases, escalators and elevators. Given that the dataset is based on natural sequences, there are much more training samples for walking than for the other modes.

The entire dataset has been labeled with the different motion modes. The labels are not used in training, only in visualization of the results. Since it is hard to strictly define the modes, they were manually annotated using subjective analysis of the video. The result still illuminates on the performance and shortcomings.

The histograms of the instantaneous speed of each of the classes can be seen in Figure~\ref{fig:histograms}. The distribution of maximum and minimum speeds shows that there is a lot of variation on each frame, but the mean remains relatively constrained.

\begin{figure}[!t]
  \centering\footnotesize
  \pgfplotsset{yticklabel style={rotate=90}, ylabel style={yshift=-15pt},clip=true,scale only axis,axis on top,clip marker paths=true}
  \setlength{\figurewidth}{.4\textwidth}
  \setlength{\figureheight}{.65\figurewidth}  
  \input{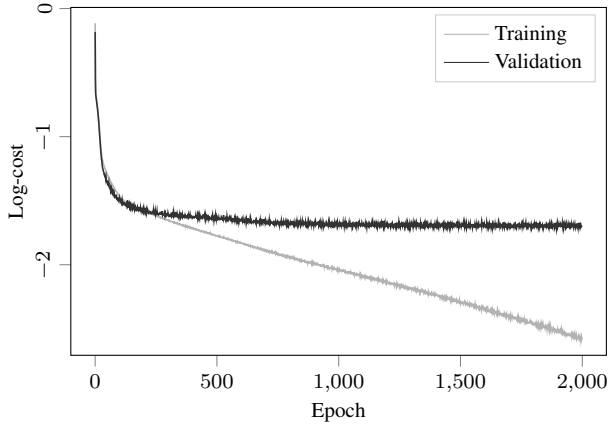}
  \caption{Log-cost progression by epoch during training for one of the folds. As the training progresses the cost stabilizes for the validation (held-out) set, indicating convergence.}
  \label{fig:test}
\end{figure}

\begin{figure}[t]
  \centering\footnotesize
  \pgfplotsset{yticklabel style={rotate=90}, ylabel style={yshift=-15pt},clip=true,scale only axis,axis on top,clip marker paths=true}
  \setlength{\figurewidth}{.4\textwidth}
  \setlength{\figureheight}{\figurewidth}  
  \input{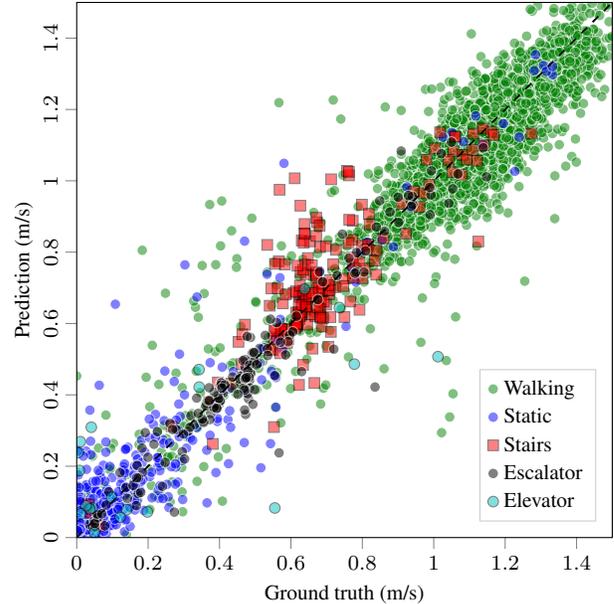}
  \caption{Labeled scattered validation results for 10-fold cross-validation. Ideally all cases would fall on the identity line. The labels were not used during training; they are just for interpreting the results.}
  \label{fig:scatter}
\end{figure}

\begin{figure}[t]
  \centering\footnotesize
  \pgfplotsset{yticklabel style={rotate=90}, ylabel style={yshift=-15pt},clip=true,scale only axis,axis on top,clip marker paths=true}
  \setlength{\figurewidth}{.4\textwidth}
  \setlength{\figureheight}{.65\figurewidth}  
  \input{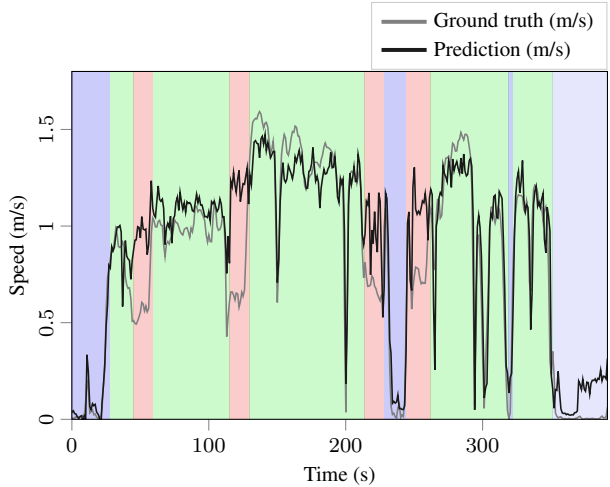}
  \caption{Speeds from an independent test sequence featuring movement on three floor levels. The background color indicates the motion mode (\textcolor{mycolor11}{\textbf{static}}, \textcolor{mycolor12}{\textbf{walking}}, and \textcolor{mycolor10}{\textbf{stairs}}).}
  \label{fig:test}
\end{figure}

\section{Experiments}
\label{sec:experiments}

We present experimental validation primarily for estimating the speed, but also show proof-of-concept results for using the speed estimates for improving odometry estimates.

\subsection{Network performance}
The CNN network was validated with 10-fold cross-validation, the average RMSE on the validation data is 0.13~m/s. The result of the 10-fold cross-validation is plotted against the ground truth in Figure~\ref{fig:scatter}.

Green dots show samples labeled as walking. Walking motion is the bulk of the training data. Blue dots show the samples marked as static. Static is defined as not in the process of going from one place to another---this includes standing and short `looking around' movements. For these two modes the predictions agree well with the ground truth. During standstill slight leaking towards higher speeds can be noticed. Red squares show samples labeled as stairs. Stairs samples include the short walking samples in between staircases. Dark gray dots show samples labeled as escalator. Just as with stairs, these include the short walks between escalators. Finally, cyan dots show the few elevator examples, these were recorded in a glass-walled elevator in order to keep the functionality of the ARKit.

The network was also tested with a newly recorded sequence on the same device that contains walking, standing still, and going up and down stairs. The whole ground truth speed and prediction by the network can be seen in Figure~\ref{fig:test}. The sequence is sampled every second, so the windows have a 50\% overlap. The resulting  RMSE  is 0.20~m/s. The background color shows the label for that portion of the test sequence. The colors are the same used in Figure~\ref{fig:scatter}.

\begin{figure*}[t]
  \centering\tiny
  \pgfplotsset{yticklabel style={rotate=90}, ylabel style={yshift=-15pt},clip=true,scale only axis,axis on top,clip marker paths=true,major grid style={dotted,black!25}}
  \setlength{\figurewidth}{.14\textwidth}
  \setlength{\figureheight}{2.0\figurewidth}
  \begin{subfigure}[b]{.15\textwidth}
    \input{./fig/path-no-pseudo.tex}
    \caption{No constraints}
  \end{subfigure}
  \hspace*{\fill}
  \begin{subfigure}[b]{.15\textwidth}
    \input{./fig/path-pseudo.tex}
    \caption{Pseudo-speed}
  \end{subfigure}
  \hspace*{\fill}
  \begin{subfigure}[b]{.15\textwidth}
    \input{./fig/path-cnn.tex}
    \caption{CNN constraint}
  \end{subfigure}
  \hspace*{\fill}
  \begin{subfigure}[b]{\columnwidth}
    {\includegraphics[width=.9\columnwidth,trim=100 100 90 90,clip]{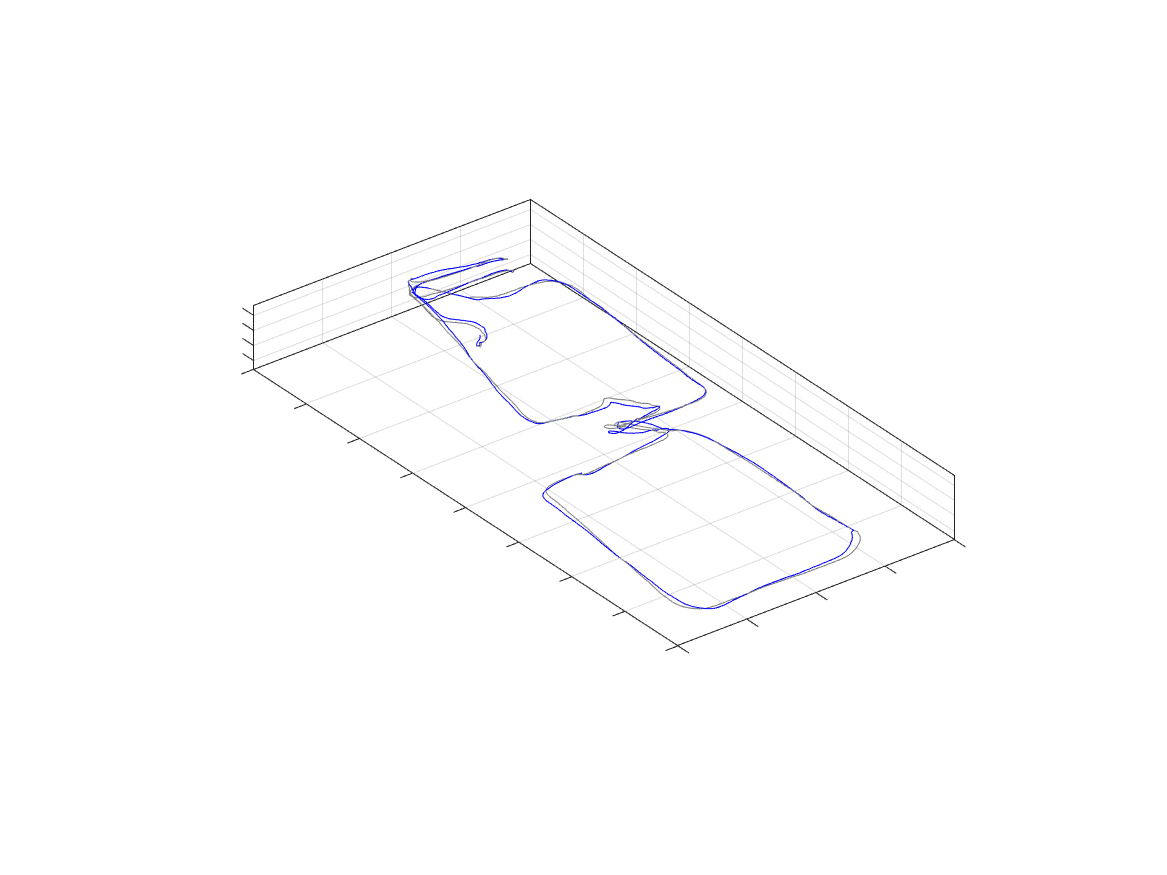}}
    \caption{Perspective view of movement on three floors}
  \end{subfigure}  
  \caption{The reconstructed metric path from the test sequence in Fig.~\ref{fig:test}. (a)--(c) show the INS results with (a)~no speed constraint, (b)~a pseudo-constraint, and (c)~using the CNN estimated speed. The ARKit ground truth is in gray and fix points shown by dots. The RMSE errors are 238.38~m, 0.87~m, and 0.62~m, respectively.}
  \label{fig:sequence}
\end{figure*}

\subsection{Inertial navigation with speed constraints}
\label{sec:ins}
The relevance of the momentary speed in INS systems is put to the test in a simple experiment. The test dataset used for Figure~\ref{fig:sequence} is put trough an INS based fix-point interpolator where the fix-points are sampled from the position data every 17~seconds. The INS is based on the system presented in \cite{Solin+Cortes+Rahtu+Kannala}.

The original system uses zero velocity updates and a constant pseudo-speed measurement to prevent fast drifts that are common in these navigation systems. In this case the pseudo-measurement is replaced with the regressed speed and the measurement noise is reduced to more accurately reflect the knowledge gained from the CNN. The pseudo-speed measurement is of the form  $h_\text{pseudo}(\vect{x}) = \norm{\vect{v}}$, where $\vect{v}$ is the speed component of the state vector $\vect{x}$.

The original INS system uses constant 0.75~m/s speed updates on the L2 norm of the velocity vector  as it is stored on the extended Kalman Filter (EKF) state. This maintains the speed at a constrained value. The updated version uses the regressed speed as the pseudo-speed measurement. This is important when the actual speed differs from the 0.75~m/s the most. In this case the measurement is more certain below 0.2~m/s.

Figure~\ref{fig:sequence} shows the result of the test sequence reconstruction using the INS based interpolation. In (d) the whole path is shown, it spans three floors of an office building. Subfigure (a) shows the path interpolated using no speed constrains at all, (b) shows the path interpolated using the constant pseudo-speed (as in \cite{Solin+Cortes+Rahtu+Kannala}), and (c) shows the path interpolated using our CNN regressed speed.

The RMSE error for the reconstructed paths are: \\
\vspace*{-1em}
\begin{center}
\begin{tabular}{rl}
  No contraints: & 238.38 m \\
  Pseudo-speed: & 0.87 m \\
  CNN constraint: & 0.62 m
\end{tabular}
\end{center}

From both the RMSEs and Figure~\ref{fig:sequence} it is clear that the unconstrained estimation scheme diverges, while the speed-contrained ones work rather well. In this case, we gain a clear improvement in terms of RMSE by constraining the system by the estimated  speed.

\section{Discussion}

The scatter plot of labeled data in Figure~\ref{fig:scatter} presents several characteristics of the data. The walking motion is clearly over represented in the training data, however, we posit that walking is the main mode of medium range transportation where smartphone  based INS is used. The network performs relatively well all around, even with the small sample size for some modes.
The test sequence shows that the speed prediction is very accurate for walking and standing still. The network regresses the speed and has no problem in the transition between modes. However, there are errors in all the staircase sequences. It is probably due to under representation in the training data.

The errors are mostly in small portions of one mode surrounded by other modes, for example a three step stair where there was walking before and after. Most people have a different stride to tackle this kind of obstacle as opposed to a full staircase.
The histograms in Figure~\ref{fig:histograms} show how the instantaneous speed is distributed among the different motion modes. Even though the instantaneous speeds are not very compact, the mean speed is fairly consistent within each mode. 

Section~\ref{sec:ins} shows how the regressed speed helps a basic INS system by adjusting the pseudo-update drift with the information from the CNN. The low speed pseudo-update covers the space between the ZUPTs. The result can be seen in the first few seconds of the capture where the motion is minimal.

\section{Conclusion}
In this paper we have proposed a scheme for free three-dimensional inertial navigation, that combines classical strict physics-driven inertial navigation with a purely data-driven approach of injecting additional knowledge through estimating the momentary speed by a CNN. We see this as a good split between more strict model-based modelling and `blind' learning from data. 

Our main interest was in evaluating the CNN model. We present an approach to regress the momentary speed based on a two second window of IMU data. The data was trained on the displacement as reported by the visual-inertial method ARKit.

The system was tested on a new sequence unknown to the training. The results were accurate on both walking and standing still modes, but could be improved for use on stairs. Finally, in a proof-of-concept study, we used the regressed speed in a functioning INS system to help constrain the movement. We gained improvements in terms of RMSE.

For the speed estimates to be more helpful, there are a number of possible future research directions to consider. In free use cases, kicking off the estimation can be a challenge. Furthermore, considering the speeds in the horizontal and vertical directions separately might help the challenges related to movement in escalators and stairs. How well the methods generalize over devices with different sensor biases should also be tackled by broader sets of training data.

The data and codes for the speed regression problem can be accessed at: \url{https://aaltovision.github.io/deep-speed-constrained-ins/}.

\paragraph*{Acknowledgements.} 
This research was supported by the Academy of Finland grants 308640, 277685, and 295081. We acknowledge the computational resources provided by the Aalto Science-IT project.

\section{References}
\balance
\begingroup
\renewcommand{\section}[1]{}
\small
\bibliographystyle{IEEEbib}
\bibliography{bibliography}

\begin{thebibliography}{10}

\bibitem{Jekeli:2001}
Christopher Jekeli,
\newblock {\em Inertial Navigation Systems with Geodetic Applications},
\newblock Walter de Gruyter, Berlin, Germany, 2001.

\bibitem{Britting:2010}
Kenneth~R Britting,
\newblock {\em Inertial Navigation Systems Analysis},
\newblock Wiley-Interscience, New York, 2010.

\bibitem{Solin+Cortes+Rahtu+Kannala}
Arno Solin, Santiago Cortes, Esa Rahtu, and Juho Kannala,
\newblock ``Inertial odometry on handheld smartphones,''
\newblock in {\em Proceedings of the International Conference on Information
  Fusion (FUSION)}, 2018.

\bibitem{Yan+Shan+Furukawa:2017}
Hang Yan, Qi~Shan, and Yasutaka Furukawa,
\newblock ``{RIDI}: {R}obust {IMU} double integration,''
\newblock {\em arXiv preprint arXiv:1712.09004}, 2017.

\bibitem{Chen+Lu+Markham+Trigoni:2018}
Changhao Chen, Xiaoxuan Lu, Andrew Markham, and Niki Trigoni,
\newblock ``{IONet}: {L}earning to cure the curse of drift in inertial
  odometry,''
\newblock in {\em Proceedings of AAAI Conference on Artificial Intelligence},
  2018.

\bibitem{Sarkka+Tolvanen+Kannala+Rahtu:2015}
Simo S\"arkk\"a, Ville Tolvanen, Juho Kannala, and Esa Rahtu,
\newblock ``Adaptive {K}alman filtering and smoothing for gravitation tracking
  in mobile systems,''
\newblock in {\em Proceedings of the International Conference on Indoor
  Positioning and Indoor Navigation (IPIN)}, Banff, Canada, 2015, pp. 1--7.

\bibitem{Foxlin:2005}
Eric Foxlin,
\newblock ``Pedestrian tracking with shoe-mounted inertial sensors,''
\newblock {\em Computer Graphics and Applications}, vol. 25, no. 6, pp. 38--46,
  2005.

\bibitem{Nilsson+Gupta+Handel:2014}
John-Olof Nilsson, Amit~K. Gupta, and Peter H\"andel,
\newblock ``Foot-mounted inertial navigation made easy,''
\newblock in {\em Proceedings of the International Conference on Indoor
  Positioning and Indoor Navigation (IPIN)}, Busan, Korea, 2014, pp. 24--29.

\bibitem{Solin+Cortes+Rahtu+Kannala:2018}
Arno Solin, Santiago Cortes, Esa Rahtu, and Juho Kannala,
\newblock ``{PIVO}: {P}robabilistic inertial-visual odometry for
  occlusion-robust navigation,''
\newblock in {\em Proceedings of the IEEE Winter Conference on Applications of
  Computer Vision (WACV)}, 2018.

\bibitem{maplab}
Thomas Schneider, Marcin Dymczyk, Marius Fehr, Kevin Egger, Simon Lynen, Igor
  Gilitschenski, and Roland Siegwart,
\newblock ``maplab: An open framework for research in visual-inertial mapping
  and localization,''
\newblock {\em IEEE Robotics and Automation Letters}, 2018.

\bibitem{Harle:2013}
Robert Harle,
\newblock ``A survey of indoor inertial positioning systems for pedestrians,''
\newblock {\em Communications Surveys \& Tutorials}, vol. 15, no. 3, pp.
  1281--1293, 2013.

\bibitem{Xiao+Wen+Markham+Trigoni:2014}
Zhuoling Xiao, Hongkai Wen, Andrew Markham, and Niki Trigoni,
\newblock ``Robust pedestrian dead reckoning ({R-PDR}) for arbitrary mobile
  device placement,''
\newblock in {\em Proceedings of the International Conference on Indoor
  Positioning and Indoor Navigation (IPIN)}, Busan, Korea, 2014, pp. 187--196.

\bibitem{Mansur+Makihara+Aqmar+Yagi:2014}
Al~Mansur, Yasushi Makihara, Rasyid Aqmar, and Yasushi Yagi,
\newblock ``Gait recognition under speed transition,''
\newblock in {\em Proceedings of the IEEE Conference on Computer Vision and
  Pattern Recognition}, 2014, pp. 2521--2528.

\bibitem{LeCun+Boser:1989}
Yann~A LeCun, Bernhard~E Boser, John~S Denker, Donnie Henderson, R.E. Howard,
  Wayne~E Hubbard, and Lawrence~D Jackel,
\newblock ``Backpropagation applied to handwritten zip code recognition,''
\newblock {\em Neural Computation}, vol. 1, no. 4, pp. 541--551, 1989.

\bibitem{Kingma+Ba:2014}
Diederik~P. Kingma and Jimmy Ba,
\newblock ``Adam: {A} method for stochastic optimization,''
\newblock in {\em Proceedings of the 3rd International Conference for Learning
  Representations (ICLR)}, San Diego, CA, 2015.

\bibitem{Cortes+Solin+Rahtu+Kannala:2018-ADVIO}
Santiago Cort\'es, Arno Solin, Esa Rahtu, and Juho Kannala,
\newblock ``{ADVIO}: {A}n authentic dataset for visual-inertial odometry,''
\newblock in {\em Proceedings of the European Conference on Computer Vision
  (ECCV)}, Munich, Germany, 2018.

\end{thebibliography}

\endgroup

\end{document}